\documentclass[journal,letterpaper,twoside,10 pt]{IEEEtran}

\usepackage{cite}
\usepackage[colorlinks = true, urlcolor = blue,citecolor = black, linkcolor = black]{hyperref}
\usepackage{balance}
\usepackage[font = footnotesize]{caption}
\usepackage{amssymb}
\usepackage{mathtools}
\usepackage{amsmath}
\usepackage{url}
\usepackage{optidef}

\DeclareMathOperator*{\argmax}{arg\,max}

\begin{document}

\title{Safety-Aware Perception for Autonomous Collision Avoidance in Dynamic Environments}

\author{Ryan M. Bena$^{1}$, Chongbo Zhao$^{1}$, and Quan Nguyen$^{1}$%
\thanks{$^{1}$The authors are with the Department of Aerospace and Mechanical Engineering, University of Southern California, Los Angeles, CA 90089, USA ({\tt bena@usc.edu; chongboz@usc.edu; quann@usc.edu}).}
}%

\maketitle

\begin{abstract}

Autonomous collision avoidance requires accurate environmental perception; however, flight systems often possess limited sensing capabilities with field-of-view (FOV) restrictions. To navigate this challenge, we present a safety-aware approach for online determination of the optimal sensor-pointing direction $\psi_\text{d}$ which utilizes control barrier functions (CBFs). First, we generate a spatial density function $\Phi$ which leverages CBF constraints to map the collision risk of all local coordinates. Then, we convolve $\Phi$ with an attitude-dependent sensor FOV quality function to produce the objective function $\Gamma$ which quantifies the total observed risk for a given pointing direction. Finally, by finding the global optimizer for $\Gamma$, we identify the value of $\psi_\text{d}$ which maximizes the perception of risk within the FOV. We incorporate $\psi_\text{d}$ into a safety-critical flight architecture and conduct a numerical analysis using multiple simulated mission profiles. Our algorithm achieves a success rate of $88-96\%$, constituting a $16-29\%$ improvement compared to the best heuristic methods. We demonstrate the functionality of our approach via a flight demonstration using the Crazyflie 2.1 micro-quadrotor. Without \textit{a priori} obstacle knowledge, the quadrotor follows a dynamic flight path while simultaneously calculating and tracking $\psi_\text{d}$ to perceive and avoid two static obstacles with an average computation time of 371 $\mu$s.
\end{abstract}

\section{INTRODUCTION}
\label{Section01}

\IEEEPARstart{C}{ollision} avoidance is one of the primary safety-critical control tasks associated with autonomous systems, particularly \textit{unmanned aerial vehicles} (UAVs). Whether UAVs are autonomously tracking a flight path through an unknown environment, surveilling a target during reconnaissance operations, or exploring an extraterrestrial location for scientific research, it is always paramount, for the continuing safe operation of the UAV, to avoid collisions with both static and dynamic obstacles. This challenge can be addressed through the use of \textit{control barrier function} (CBF) based \textit{quadratic programs} (QPs). Introduced in \cite{ames2014cbf} and expounded in \cite{ames2016cbf,nguyen2016exponential}, CBF-QPs provide a computationally-efficient and minimally-invasive means of enforcing safety-critical control objectives on a system. The application of CBF-QPs to quadrotor UAV collision avoidance has been demonstrated in \cite{wu2016safety,wu2016safetycritical,doeser2020invariant}.

\begin{figure}[t]
\centering
\medskip
\includegraphics[width=\linewidth]{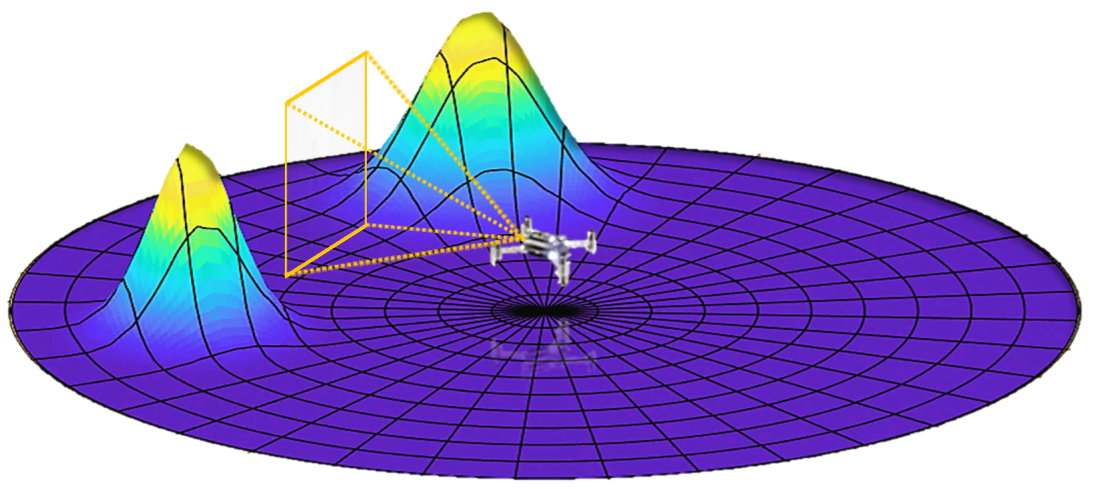}
\caption{\textbf{Safety-Aware Perception.} Our optimal safety-aware perception methodology. Using an onboard sensor with limited FOV, the UAV determines its ideal sensor-pointing direction based on a local spatial density function. The peaks of the density function represent risk levels in the local environment, calculated using CBFs.}
\label{CoverPhoto}
\vspace{-4ex}
\end{figure}

CBF-based safety-critical flight controllers require perception of obstacles in the local environment, yet small autonomous flight systems are generally equipped with basic sensor suites which include: gyroscopes, accelerometers, and, less frequently, magnetometers and GPS receivers. These sensors allow UAVs to maintain attitude stability while performing simple navigational tasks. On the other hand, it is less common for UAVs to be equipped with optical cameras, LiDAR devices, time-of-flight sensors, etc., especially as vehicle scales decrease, because these environmental perception sensors consume scarce payload and computing resources. As a result, micro-UAV researchers are often forced to rely on \textit{a priori} obstacle knowledge to map static environments \cite{xu2018safe,freire2021flatness} and off-board motion capture systems to monitor dynamic environments \cite{wang2017safe}. These imperfect solutions motivate the need for obstacle detection and collision avoidance methodologies which consider practical sensing and processing limitations. To this effect, we establish a common sensing scenario involving a quadrotor UAV equipped with a single onboard sensor characterized by limited range and a body-fixed \textit{field-of-view} (FOV). In this scenario, optimally pointing the onboard sensor becomes a coverage control problem.

\begin{figure*}[t]
\begin{center}
\medskip
\includegraphics[width=\linewidth]{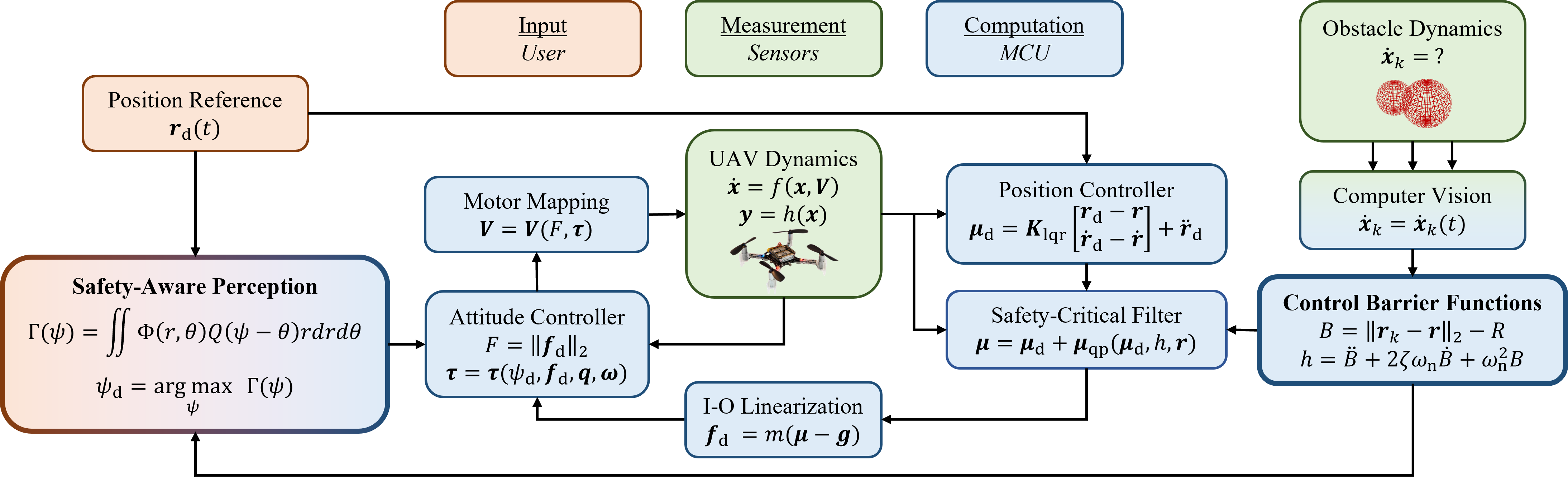}
\caption{\textbf{Flight Control Block Diagram.} The structure of the safety-critical controller for our UAV. Onboard sensors gather UAV and dynamic obstacle state information (green), and provide it to a feedback control algorithm (blue). The controller uses CBFs in a safety-critical QP to avoid collisions while tracking a user-defined position reference $\boldsymbol{r}_\text{d}$ (orange). With our safety-aware perception algorithm, the yaw reference $\psi_\text{d}$ is used to orient a fixed onboard sensor/camera to enable obstacle detection and tracking.
\label{UAVControlDiagram}}
\end{center}
\vspace{-4ex}
\end{figure*}

Coverage control describes a class of perception control problems which aim to find desirable sensor positions/orientations given environmental parameters and performance objectives. Optimal coverage control techniques often involve the coordination of one or more sensors to maximize spatial data observations \cite{hubel2008coverage,reddy2012camera,forstenhaeusler2015experimental,arslan2018coverage}, and in recent years these methods have been extensively applied to UAVs for ground surveillance and exploration tasks \cite{paull2014sensor,mebarki2015nonlinear,funada2019visual,renzaglia2020common,dan2021persistent,zhang2021robust}. A few research teams have also previously tackled the limited-perception servoing problem for obstacle detection and avoidance. In \cite{finean2022where,ryu2022confidence} novel heading control methodologies are presented for mobile ground-based robots in unknown dynamic environment, and in \cite{murali2019perception,spasojevic2020perception} offline planning approaches are used to calculate optimal quadrotor UAV yaw trajectories for \textit{a priori} known static flight spaces. The research in \cite{li2023robocentric} uses a \textit{model predictive control} (MPC) method for online visual servoing, but the method does not extend to general navigation through dynamic multi-obstacle environments. This is often the case for UAV-based aerial object tracking and avoidance algorithms, such as those presented in \cite{penin2018vision,sheng2019image,liu2021visual,qin2023perception,kaufmann2023champion}. Another drawback in the literature is a lack of efficient approaches for small-scale systems. The methods in \cite{chen2016online,kong2021avoiding} populate and maintain dense environmental maps which are used to run computationally intensive planning algorithms. The PANTHER algorithm \cite{tordesillas2022panther,tordesillas2023deep} addresses the problem of perception in dynamic environments but requires significant onboard computation, and it differs in its approach towards quantifying safety and collision risk. Perhaps the most efficient state-of-the-art approach, the \textit{active sense and avoid} (ASAA) method \cite{chen2021active}, computes a sensor-pointing direction to minimize a cost function which considers reference trajectory, velocity direction, obstacle locations, general exploration, and control effort. However, this approach differs from ours in two key aspects: 1) the ASAA method is intended for implementation via a maneuverable camera, not the UAVs yaw \textit{degree-of-freedom} (DOF), and 2) the ASAA method does not leverage CBFs, neither for perception nor collision avoidance.

In this paper, we present a safety-aware perception algorithm which calculates the optimal sensor FOV for obstacle detection and avoidance in dynamic environments. Our approach, which merges CBF-based safety-critical control and coverage control concepts to facilitate collision-free navigation, is presented via the following structure: Section \ref{Section02} provides a synopsis of our UAV control architecture; Section\;\ref{Section03} presents our novel environmental perception algorithm, Section\;\ref{Section04} depicts numerical results from a series of UAV simulations, highlighting the benefits of our technique over several heuristic methods, Section\;\ref{Section05} demonstrates the successful implementation of our algorithm on a Crazyflie 2.1 micro-quadrotor, and Section\;\ref{Section06} concludes the manuscript by summarizing the findings and presenting plans for future research. 

\section{UAV SAFETY-CRITICAL CONTROL}
\label{Section02}

The function of the flight controller is to compel the UAV to track a position reference trajectory $\boldsymbol{r}_\text{d}\in\mathbb{R}^3$ and an independent yaw reference trajectory $\psi_\text{d}\in\mathbb{R}$ while simultaneously avoiding collisions with dynamic obstacles. This control architecture uses, as its foundation, an evolved version of the UAV tracking control scheme presented in \cite{bena2022position,bena2023performance}. To the nominal tracking controller we add a safety-critical filter and an online safety-aware perception algorithm, producing the final structure shown in Fig.\;\ref{UAVControlDiagram}. 

To follow a dynamic position reference, we estimate the position of the UAV $\boldsymbol{r}\in\mathbb{R}^3$ and its time derivative $\dot{\boldsymbol{r}}$ and employ state feedback, in the form of a \textit{linear quadratic regulator} (LQR) algorithm, with an additional term for tracking. The resultant position controller is

\begin{equation}
    \boldsymbol{\mu}_\text{d} = \boldsymbol{K}_\text{lqr}
    \left[
    \begin{array}{c}
         \boldsymbol{r}_\text{d} - \boldsymbol{r}  \\
         \dot{\boldsymbol{r}}_\text{d} - \dot{\boldsymbol{r}}
    \end{array}
    \right]
    + \ddot{\boldsymbol{r}}_\text{d},
\label{LQR}
\end{equation}

\noindent where $\boldsymbol{\mu}_\text{d}\in\mathbb{R}^3$ is the desired virtual acceleration representing the ideal acceleration of the UAV to track $\boldsymbol{r}_\text{d}$, and $\boldsymbol{K}_\text{lqr}\in\mathbb{R}^{3\times6}$ is the LQR-based state feedback gain matrix. 

The desired control $\boldsymbol{\mu}_\text{d}$ is sent to a QP which modifies the final virtual acceleration $\boldsymbol{\mu}\in\mathbb{R}^3$ to ensure convergence to a safe-set $\mathcal{S}$ defined as

\begin{equation}
    \mathcal{S} = \{\boldsymbol{r}\in\mathbb{R}^3: B(\boldsymbol{r}) \geq 0\},
\label{SafeSet}
\end{equation}

\noindent where $B(\boldsymbol{r}):\mathbb{R}^3 \mapsto \mathbb{R}$ defines the boundary of the safe-set. For spherical obstacles,

\begin{equation}
    B(\boldsymbol{r}) = ||\boldsymbol{r}_\text{c}-\boldsymbol{r}||_2 - R,
\label{SphericalSafeSet}
\end{equation}

\noindent where $\boldsymbol{r}_\text{c}\in\mathbb{R}^3$ is the obstacle center position and $R\in\mathbb{R}$ is the obstacle barrier radius, which accounts for UAV size. 

The inequality in (\ref{SafeSet}) defines the condition for which a position is considered safe. CBFs provide a method by which convergence of $\boldsymbol{r}$ to $\mathcal{S}$ and forward invariance of $\mathcal{S}$ can be guaranteed by satisfying a constraint on the control input \cite{ames2019cbf}. The particular safe-set defined by (\ref{SphericalSafeSet}) can be employed as a CBF of relative degree 2, so we need an exponential CBF constraint defined by

\begin{equation}
    h(\boldsymbol{r},\boldsymbol{\mu}) = \ddot{B}(\boldsymbol{r},\boldsymbol{\mu}) + 2\zeta\omega_\text{n} \dot{B}(\boldsymbol{r})+\omega_\text{n}^2 B(\boldsymbol{r}),
\label{CBF}
\end{equation}

\noindent where, for a given translational state, forward invariance of $\mathcal{S}$ can be enforced via the inequality
\begin{equation}
    h(\boldsymbol{\mu}) \geq 0.
    \label{CBFConstraint}
\end{equation}

\noindent The parameters $\zeta\in\mathbb{R}$ and $\omega_\text{n}\in\mathbb{R}$ must be chosen such that the roots of the characteristic equation of (\ref{CBF}) are negative real \cite{nguyen2016exponential}. This holds for $\zeta\geq1$ and $\omega_\text{n}>0$. 

The inequality in (\ref{CBFConstraint}) is linear with respect to $\boldsymbol{\mu}$ and can therefore be rearranged into the form

\begin{equation}
    \boldsymbol{A}_\text{cbf} \boldsymbol{\mu} \leq b_\text{cbf},
\label{LinearCBFConstraint}
\end{equation}

\noindent where $\boldsymbol{A}_\text{cbf}\in\mathbb{R}^{1\times 3}$ and $b_\text{cbf}\in\mathbb{R}$ represent the parameters of the final linear inequality constraint for a particular obstacle. 

Constraints for every obstacle are enforced on the control input using quadratic programming. Given the desired control action $\boldsymbol{\mu}_\text{d}$ generated using (\ref{LQR}), the goal of CBF-based safety-critical control is to determine the minimum deviation $\boldsymbol{\mu}_\text{qp}\in\mathbb{R}^3$ which, when added to $\boldsymbol{\mu}_\text{d}$, produces a final control, 

\begin{equation}
    \boldsymbol{\mu}=\boldsymbol{\mu}_\text{d}+\boldsymbol{\mu}_\text{qp}, 
\end{equation}

\noindent that satisfies (\ref{LinearCBFConstraint}). Furthermore, the final control $\boldsymbol{\mu}$ should be physically achievable by the plant with minimal destabilizing effect on the tracking control task. This combination of requirements can be expressed via a QP of the form:

\noindent\rule{\linewidth}{0.5pt}
\noindent \textbf{CLF-CBF-QP}
\begin{argmini}|l|
    {\boldsymbol{\mu}_\text{qp}}{\frac{1}{2}\boldsymbol{\mu}_\text{qp}^T\boldsymbol{H}_\text{qp}\boldsymbol{\mu}_\text{qp} + \frac{1}{2}\xi\delta^2}{}{}    \addConstraint{\boldsymbol{A}_\text{cbf} \boldsymbol{\mu}}{\leq b_\text{cbf}}\addConstraint{\boldsymbol{A}_\text{clf}\boldsymbol{\mu}}{\leq b_\text{clf}+\delta}    \addConstraint{\boldsymbol{\mu}_\text{min}}{\leq \boldsymbol{\mu}}{\leq \boldsymbol{\mu}_\text{max}}\addConstraint{\delta}{\geq 0.}
\label{QPFilter}
\end{argmini}
\noindent\rule{\linewidth}{0.5pt}

\noindent Here, $\boldsymbol{H}_\text{qp}\in\mathbb{R}^{3\times 3}$ is a positive diagonal matrix. The \textit{control Lyapunov function} (CLF) constraint, defined by $\boldsymbol{A}_\text{clf}\in\mathbb{R}^{1\times 3}$ and $b_\text{clf}\in\mathbb{R}$, is derived from a valid CLF \cite{ames2014rapidly}. In particular, we choose a quadratic CLF to match the exponential convergence of (\ref{LQR}). Bounds on the control input are captured in $\boldsymbol{\mu}_\text{min}\in\mathbb{R}^{3}$ and $\boldsymbol{\mu}_\text{max}\in\mathbb{R}^{3}$ based on UAV acceleration limits. The slack variable $\delta\in\mathbb{R}$ and its cost weighting factor $\xi\in\mathbb{R}$ soften the CLF constraint to increase feasibility of the QP.

Using \textit{input-output} (I-O) linearization \cite{isidori1985nonlinear}, the desired control force vector $\boldsymbol{f}_\text{d}\in\mathbb{R}^3$ is computed via

\begin{equation}
    \boldsymbol{f}_\text{d} = m\left(\boldsymbol{\mu}-\boldsymbol{g}\right),
\end{equation}

\noindent where $m\in\mathbb{R}$ is the mass of the quadrotor UAV, and $\boldsymbol{g}=[0\;\;0\;\;-9.81]^T\;\text{m}/\text{s}^2$ is Earth's gravitational acceleration. 

By design, most quadrotor UAVs are restricted to applying thrust in the body-vertical direction. Therefore, to achieve the desired control $\boldsymbol{f}_\text{d}$ the body-vertical axis must be aligned with the desired force vector direction using an attitude controller. First, combining the force vector with the independently-defined yaw reference $\psi_\text{d}$, we compose a desired attitude quaternion $\boldsymbol{q}_\text{d}\in\mathbb{R}^4$. Generating the optimal value of $\psi_\text{d}$ for safety-aware perception is the subject of Section\;\ref{Section03}.

As with the position controller, we employ state feedback with tracking and I-O linearization to produce:

\begin{equation}
    \boldsymbol{q}_\text{e} = \left[
    \begin{array}{c}
         m_\text{e} \\
         \boldsymbol{n}_\text{e} 
    \end{array}
    \right] = \boldsymbol{q}^{-1} \otimes \boldsymbol{q}_\text{d},
\label{QuaternionError}
\end{equation}
\begin{equation}
    \boldsymbol{\nu} = k_q \boldsymbol{n}_\text{e} \text{sgn}(m_\text{e}) + k_\omega(\boldsymbol{\omega}_\text{d} - \boldsymbol{\omega}) + \dot{\boldsymbol{\omega}}_\text{d},
\label{AttitudeController}
\end{equation}
\begin{equation}
    \boldsymbol{\tau} = \boldsymbol{J}\boldsymbol{\nu} + \boldsymbol{\omega} \times \boldsymbol{J}\boldsymbol{\omega},
 \label{AttitudeIO}   
\end{equation}

\noindent where $\boldsymbol{q}_\text{e}\in\mathbb{R}^4$ in (\ref{QuaternionError}) is the quaternion representing the desired attitude with respect to the current attitude, $\boldsymbol{q}\in\mathbb{R}^4$ is the current attitude quaternion, and $\otimes$ is the quaternion multiplication operator. In (\ref{AttitudeController}), $\boldsymbol{\nu}\in\mathbb{R}^3$ represents the control angular acceleration, $k_q\in\mathbb{R}$ and $k_\omega\in\mathbb{R}$ are positive feedback gains, $\boldsymbol{\omega}_\text{d}\in\mathbb{R}^3$ is the desired angular velocity, and $\boldsymbol{\omega}\in\mathbb{R}^3$ is the current angular velocity. In (\ref{AttitudeIO}), $\boldsymbol{\tau}\in\mathbb{R}^3$ is the attitude control torque, and $\boldsymbol{J}\in\mathbb{R}^{3\times3}$ is the UAV inertia matrix. The stability properties of this control law are detailed in \cite{bena2022position}. The torque $\boldsymbol{\tau}$ is mapped to the four actuators of the quadrotor to complete the control loop.

At this point, we return to the problem of selecting the yaw reference $\psi_\text{d}$. Because the yaw DOF is dynamically decoupled from UAV translational motion, this DOF can be used to manipulate onboard sensors with minimal impact on flight behavior, thus enabling perception of, and navigation through, dynamic environments. Our online safety-based perception methodology, which we use to compute an optimal value for $\psi_\text{d}$, is described in detail in the next section.

\section{SAFETY-AWARE PERCEPTION}
\label{Section03}

The safety guarantees associated with (\ref{QPFilter}), as summarized in \cite{ames2019cbf}, assume complete obstacle information (i.e. translational motion and size). However, with limited environmental sensors, this assumption is often violated. Consider, for example, a UAV which utilizes a single fixed onboard sensor, with a restricted FOV, for obstacle detection and tracking. Systems such as this are common, especially at the microrobotic scale where size, weight, and power limitations are particularly stringent. We aim to reduce the impact of these sensor limitations by establishing an online perception optimization algorithm which employs CBFs to improve safety. By computing the yaw reference angle $\psi_\text{d}$ which maximizes the amount of high-risk information contained within the sensor FOV, we increase the probability of collision-free flight in dynamic environments.

Let $\Phi(\boldsymbol{z}): \mathbb{R}^2 \mapsto \mathbb{R}$ be a positive semi-definite spatial \textit{density function} which describes the observational importance of a particular spatial coordinate $\boldsymbol{z}\in\mathbb{R}^2$. Depending on the application, various functions can be used for $\Phi$, but for safety-aware perception during collision avoidance, we introduce the following definition which emphasizes the importance of $N$ points of interest in the spatial environment:

\begin{equation}
    \Phi(\boldsymbol{z}) = \sum_{k=1}^N \frac{\alpha_k}{\beta_k||\boldsymbol{z}-\boldsymbol{z}_k||^2_2+1},
\label{DensityFunction}
\end{equation}

\noindent where $\boldsymbol{z}_k\in\mathbb{R}^2$ is the location of the $k\text{th}$ point of interest, and $\alpha_k\in\mathbb{R}$ and $\beta_k\in\mathbb{R}$ are parameters which define the peak height and dissipation rate of the density function. 

By identifying a translating reference frame $\mathcal{B}$ such that its origin is coincident with the center of mass of the UAV system, and its orientation is aligned with the inertial frame $\mathcal{N}$, we can redefine (\ref{DensityFunction}) as a relative spatial density function in polar coordinates $(r,\theta)$, with respect to $\mathcal{B}$. The resultant density function becomes

\begin{equation}
    \Phi(r,\theta) = \sum_{k=1}^N \frac{\alpha_k}{\beta_k\left[r^2 - 2rr_k\cos(\theta-\theta_k) + r_k^2\right]+1},
\label{PolarDensityFunction}
\end{equation}

\noindent where $(r_k,\theta_k)$ are the polar coordinates of the $k\text{th}$ point of interest with respect to the center of mass of the UAV. 

Again, the physical interpretation of (\ref{PolarDensityFunction}) can vary based on the particular UAV flight control application. In our safety-critical application, we primarily use the individual density function terms to characterize the collision risk of each obstacle in the local environment. To accomplish this, we reintroduce the CBF constraint definition in (\ref{CBF}). For a given obstacle, the value of its CBF constraint can be conveniently transformed into a measure of collision risk. When the value of the CBF constraint is large, the risk of a collision is low; as the CBF constraint value decreases, the collision risk increases. Leveraging this behavior, the height of the corresponding density function peak $\alpha_k$ can be related to the value of the CBF constraint by

\begin{equation}
    \alpha_k = \alpha_\text{obs} e^{-\gamma h_k},
\label{Peak}
\end{equation}

\noindent where $\alpha_\text{obs}\in\mathbb{R}$ defines the nominal peak height, $\gamma\in\mathbb{R}$ characterizes the risk decay rate, and $h_k\in\mathbb{R}$ is the current value of the obstacle CBF constraint, obtained from (\ref{CBF}).

\begin{figure}[t]
\begin{center}
\medskip
\includegraphics[width=\linewidth]{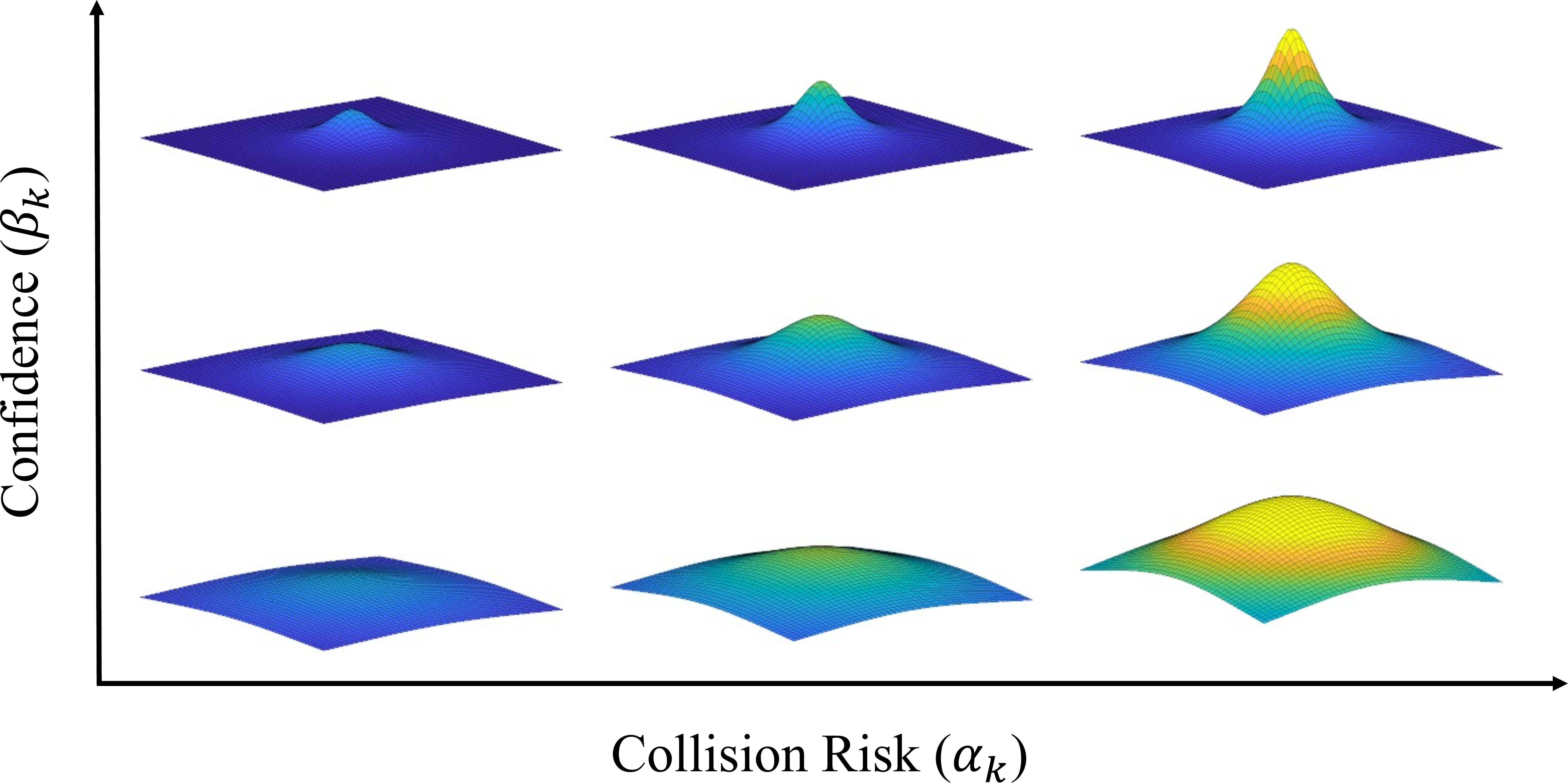}
\caption{\textbf{Density Function Behavior.} The density function $\Phi(r,\theta)$ for a particular obstacle changes with its values of collision risk $\alpha_k$ and confidence $\beta_k$. The density function peak heights modulate as a function of $\alpha_k$ while the peak widths expand and contract based on $\beta_k$.}
\label{DensityFunctions}
\end{center}
\vspace{-4ex}
\end{figure}

We employ a similar definition to quantify obstacle location confidence. When an obstacle is within the sensor FOV, the UAV has maximum confidence in the obstacle's position. Meanwhile, confidence decreases when the obstacle leaves the FOV. This behavior determines the value of $\beta_k$ by

\begin{equation}
    \beta_k = \beta_\text{obs} e^{-\lambda\tau_k},
\label{Dissipation}
\end{equation}
    
\noindent where $\beta_\text{obs}\in\mathbb{R}$ defines the nominal dissipation factor, $\lambda\in\mathbb{R}$ characterizes the confidence decay rate, and $\tau_k\in\mathbb{R}$ represents the time since the obstacle was last observed. With the definitions in (\ref{PolarDensityFunction})-(\ref{Dissipation}), the density function peak heights modulate as a function of obstacle risk while the peak widths expand and contract based on confidence in the obstacle location. These relationships are illustrated in Fig.\;\ref{DensityFunctions}.

Next, let $Q(\boldsymbol{z},\boldsymbol{p}): \mathbb{R}^4 \mapsto \mathbb{R}$ be a positive semi-definite \textit{quality function} which describes the measurement capability of a sensor. The variable $\boldsymbol{p}\in\mathbb{R}^2$ is a unit vector indicating the pointing axis of the sensor. For a single sensor with limited range and FOV, the simplest form of a quality function is a binary value, i.e., 

\begin{equation}
    Q(\boldsymbol{z},\boldsymbol{p}) = 
    \begin{cases}    
    1 \;\;\;\; \forall \boldsymbol{z} \in \mathcal{F}\\
    0 \;\;\;\; \forall \boldsymbol{z} \notin \mathcal{F}
    \end{cases},
\label{BinaryQualityFunction}
\end{equation}

\noindent where $\mathcal{F}$ is the set of all points contained within the sensor FOV. For many applications, this is sufficient to capture sensing capability. However, measurement quality often degrades at the periphery of the FOV, as is the case with optical cameras; thus it is convenient to define a quality function which accounts for this behavior via

\begin{equation}
    Q(\boldsymbol{z},\boldsymbol{p}) = 
    \begin{cases}    
    \frac{1}{1-\cos(\kappa)}\left(\frac{(\boldsymbol{z}-\boldsymbol{z}_\text{c})^T\boldsymbol{p}}{||\boldsymbol{z}-\boldsymbol{z}_\text{c}||_2}-\cos(\kappa)\right) \;\; \forall \boldsymbol{z} \in \mathcal{F}\\
    0 \;\;\;\;\;\;\;\;\;\;\;\;\;\;\;\;\;\;\;\;\;\;\;\;\;\;\;\;\;\;\;\;\;\;\;\;\;\;\;\;\;\;\;\;\;\, \forall \boldsymbol{z} \notin \mathcal{F}
    \end{cases},
\label{QualityFunction}
\end{equation}

\noindent where $\boldsymbol{z_c}\in\mathbb{R}^2$ is the location of the sensor and $\kappa\in\mathbb{R}$ is a parameter which characterizes the angular degradation of sensing quality within the FOV \cite{arslan2018coverage}.

As with (\ref{PolarDensityFunction}), the quality function can be simplified by expressing it in polar coordinates with respect to $\mathcal{B}$. Then, $\mathcal{F}$ can be defined based on the yaw angle of the UAV $\psi\in\mathbb{R}$ as \mbox{$\psi\pm\sigma$}, where $\sigma\in\mathbb{R}$ is the half-angle covered by the camera FOV. Using this definition, we rearrange the quality function with respect to $\mathcal{B}$ such that

\begin{equation}
    Q(\psi-\theta) = 
    \begin{cases}    
    \frac{\cos(\psi-\theta)-\cos(\kappa)}{1-\cos(\kappa)} \;\;\;\; \forall (\psi-\theta) \in \pm\sigma\\
    0 \;\;\;\;\;\;\;\;\;\;\;\;\;\;\;\;\;\;\;\;\;\;\;\;\; \forall (\psi-\theta) \notin \pm\sigma\\
    \end{cases}.
\label{PolarQualityFunction}
\end{equation}

With well-defined density and quality functions, we proceed to define an objective function

\begin{equation}
    \Gamma(\psi)=\int^{\pi}_{-\pi} \int^\rho_0 \Phi(r,\theta) Q(\psi-\theta) r dr d\theta,
\label{ObjectiveFunction}
\end{equation}

\noindent as the integral of the product of the density and quality functions over the local region. The radial extent of the local region $\rho\in\mathbb{R}$ is the maximum sensor range. 

This objective function quantifies the total observed risk within a particular FOV. Therefore, by maximizing its value with respect to $\psi$, we can determine the yaw angle which allows the onboard sensor(s) to observe the most local risk. Unfortunately, (\ref{ObjectiveFunction}) is non-convex; thus, finding its global optimizer is a challenging problem. While gradient-based techniques can be utilized to find local maxima, we employ a unique approach, based on the specific forms of $\Phi$ and $Q$, to find the global optimizer.

The integral in (\ref{ObjectiveFunction}) can be simplified analytically by defining the following compound variables:

\begin{equation}
\begin{split}
    a_k &= \beta_k,\\
    b_k &= -2\beta_k r_k \cos(\theta-\theta_k),\\
    c_k &= \beta_k r_k^2 + 1.\\
\end{split}
\end{equation}

\noindent Substituting the variables, along with (\ref{PolarDensityFunction}), into (\ref{ObjectiveFunction}) yields

\begin{equation}
    \Gamma(\psi)=\int^{\pi}_{-\pi} \left[ \int^\rho_0 \sum_{k=1}^N \frac{\alpha_k r}{a_k r^2 + b_k r + c_k} dr \right] Q(\psi-\theta) d\theta.
\label{ObjectiveFunction2}
\end{equation}

The interior definite integral with respect to $r$ can be solved analytically to obtain

\begin{equation}
    \Gamma(\psi) = \int^{\pi}_{-\pi}  H(\theta) Q(\psi-\theta) d\theta,
\label{ObjectiveFunction3}
\end{equation}

\noindent where the function $H$ is

\begin{multline}
    H(\theta) = \sum_{k=1}^N\frac{\alpha_k}{2\beta_k}\biggl[\ln\left(\frac{a_k \rho^2 + b_k \rho + c_k}{c_k}\right) - \\ 
    \frac{2b_k}{\sqrt{4a_k c_k - b_k^2}}\arctan\left(\frac{\rho\sqrt{4a_k c_k - b_k^2}}{2c_k + b_k\rho}\right)\biggr].
\end{multline}

\noindent Because the value of $Q$ is only nonzero over the interval \mbox{$\psi\pm\sigma$}, the final form of the objective function can be expressed as the convolution 

\begin{equation}
    \Gamma(\psi) = \int^{\psi+\sigma}_{\psi-\sigma}  H(\theta) Q(\psi-\theta) d\theta.
\label{ObjectiveFunction4}
\end{equation}

\noindent While this convolution integral has no analytical solution, numerically evaluating (\ref{ObjectiveFunction4}) is more efficient than evaluating (\ref{ObjectiveFunction}) and thus more conducive to online computation. 

Due to the lack of convexity of (\ref{ObjectiveFunction4}), the angle which globally maximizes this objective function may not vary continuously, and there is no guarantee that the solution can be tracked by the UAV attitude controller in (\ref{QuaternionError})-(\ref{AttitudeIO}). To mitigate this, we add an additional term to the objective function which disincentivizes aggressive changes in the solution by penalizing its rate of change. This is done by

\begin{equation}
    \bar{\Gamma}(\psi) = \Gamma(\psi) - \epsilon(\psi-\psi_{\text{d},0})^2,
\label{ObjectiveFunction5}
\end{equation}

\noindent where $\psi_{\text{d},0}\in\mathbb{R}$ is the solution from the previous algorithm iteration, and $\epsilon\in\mathbb{R}$ weights the penalty term. The final desired yaw angle is defined as 

\begin{equation}
    \psi_\text{d} = \argmax_\psi \bar{\Gamma}(\psi).
\label{DesiredYaw}
\end{equation}

As specified previously, the global optimizer for this non-convex optimization problem cannot be computed with traditional gradient methods. Nonetheless, because (\ref{ObjectiveFunction5}) is a function of a single variable which only exists on the interval \mbox{$[-\pi,\pi]$}, the optimizer can be found via an incremental search which is performed by numerically evaluating (\ref{ObjectiveFunction5}) over the interval, yielding an approximate global maxima.

The main advantages of this environmental risk characterization technique, and the resultant perception methodology, are twofold. First, while many UAV processors do not have the memory capacity or computational bandwidth to continually maintain/update a dense mapping of individual environmental voxels, they can easily track a limited set of key points of interest. As such, mapping risk in local flight environment via the analytical density function in (\ref{PolarDensityFunction}) significantly reduces the online computational burden. Second, the utilization of CBFs to characterize risk provides a direct link between the estimated risk of an obstacle (\ref{Peak}) and the action the UAV flight controller will take to avoid it (\ref{QPFilter}). This correlation encourages obstacle perception during the most active moments of collision avoidance, which in turn increases the probability of collision-free flight.

\section{NUMERICAL VALIDATION}
\label{Section04}

Having presented our safety-critical control architecture and novel perception strategy, we next developed a high-fidelity UAV simulation environment using Matlab Simulink. We used this environment to assess the behavior of our proposed algorithm and conduct a numerical analysis comparing its performance to several baseline heuristic approaches. 

\begin{figure*}[t]
\begin{center}
\medskip
\includegraphics[width=\linewidth]{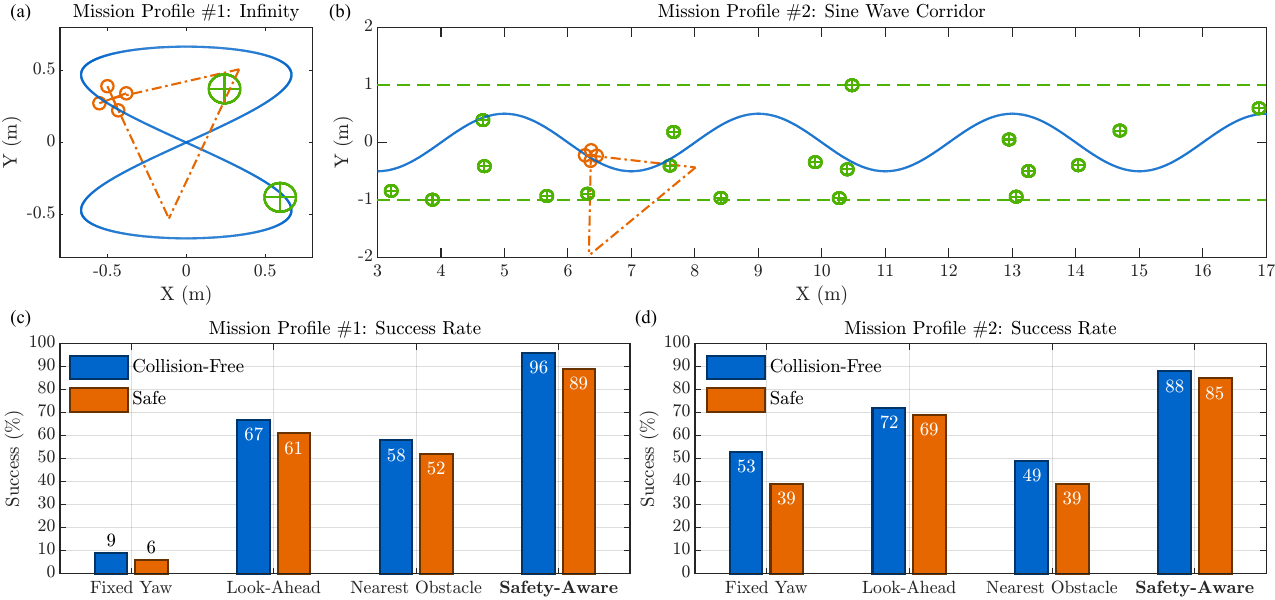}
\caption{\textbf{Dynamic Obstacle Simulations.} The mission profiles for the (a) infinity symbol track and (b) sine wave corridor which were used to evaluate various perception methodologies. The success rates for these two profiles, for 100 randomly-generated simulations, are shown in (c) and (d) respectively. Flights were categorized as \textit{collision-free} if the boundaries of the UAV never intersected those of the obstacles. Flights were further categorized as \textit{safe} if the boundaries of the UAV never crossed those of the obstacle safe-sets, including the safety factor.
\label{MonteCarlo}}
\end{center}
\vspace{-4ex}
\end{figure*}

To align our simulations with the experiments in Section\;\ref{Section05}, the simulated parameters of the UAV were chosen to closely match the Crazyflie 2.1 micro-quadrotor. We assessed our perception strategy using two flight mission profiles:

\begin{enumerate}
    \item \textit{Infinity Symbol ($\infty$) Track} - Execute several laps of an infinity symbol reference trajectory at fixed altitude with two dynamic obstacles into the local environment.
    \item \textit{Sine Wave ($\sim$) Corridor} - Follow a sinusoidal reference trajectory through a constrained $15 \times 2$ meter corridor filled with 20 dynamic obstacles. 
\end{enumerate}

We further evaluated our algorithm against three baseline heuristic approaches. The \textit{fixed yaw} method sets $\psi_\text{d}$ to a constant value without attempting
to point the sensor in a useful direction. The \textit{look-ahead} method aligns $\psi_\text{d}$ with the position reference $\boldsymbol{r}_\text{d}$. By pointing along the flight path, the UAV perceives its environment by previewing the airspace that it expects to occupy in the near future. Finally, the \textit{nearest obstacle} method points $\psi_\text{d}$ at the closest detected obstacle. If there are multiple obstacles in the environment, $\psi_\text{d}$ switches dynamically as the relative distances change.

Using randomly-generated obstacle trajectories, we conducted 100 simulated flights for each combination of mission profile and perception approach. The spherical obstacles were prescribed a radius of $10\;\text{cm}$ and a safety factor of $1.5$ which was used to define the safe-set in (\ref{SphericalSafeSet}) for the CLF-CBF-QP. To simulate physical sensor limitations, the obstacles in both scenarios could only be detected if their boundaries existed entirely within the sensor FOV. Each flight was considered \textit{collision-free} based on the true obstacle size and \textit{safe} based on the safety factor. The results from these simulations are depicted in Fig.\;\ref{MonteCarlo}, and example flight simulations are provided in the supplemental material.

For the infinity symbol mission profile shown in Fig.\;\ref{MonteCarlo}(a), our safety-aware framework produced the best results, outperforming every heuristic method with a collision-free rate of $96\%$ and a safe flight rate of $89\%$. By observing individual simulated flights, it is clear that the UAV effectively responded to the dynamic environment, smoothly transitioning between points of interest to reduce collision risk. In comparison, the fixed yaw method led to extremely poor performance. This was expected since the UAV made no effort to maneuver the sensor. The look-ahead method produced the best heuristic results, working well when obstacles were encountered along the flight path. However, safety violations occurred frequently when obstacles approached the UAV from the sides and aft. The nearest obstacle method produced marginal results because obstacles were tracked based on distance, not risk. When considering dynamic obstacles, distance is a poor measure of risk, and, in the event that all obstacle risks are low, it is more effective to simply explore the flight space. 

For the sine wave mission profile shown in Fig.\;\ref{MonteCarlo}(b), our approach again achieved the best results. While the scenario produced a slight reduction in overall success, our safety-aware perception algorithm still maintained a high standard of safe flight behavior with an $88\%$ collision-free rate. The fixed yaw and look-ahead methods both continued to exhibit substandard performance, although their success rates improved due to the unidirectional nature of the sine wave trajectory. Predictably, the performance of the nearest obstacle method degraded as a result of the dramatically increased obstacle count.

Overall, our proposed perception method exhibited a $16-29\%$ improvement over the best heuristic results. This highlights our safety-aware framework as a valuable technique merging perception with CBF-based safety-critical control.

\section{EXPERIMENTAL RESULTS}
\label{Section05}

\begin{figure}[t]
\begin{center}
\medskip
\includegraphics[width=\linewidth]{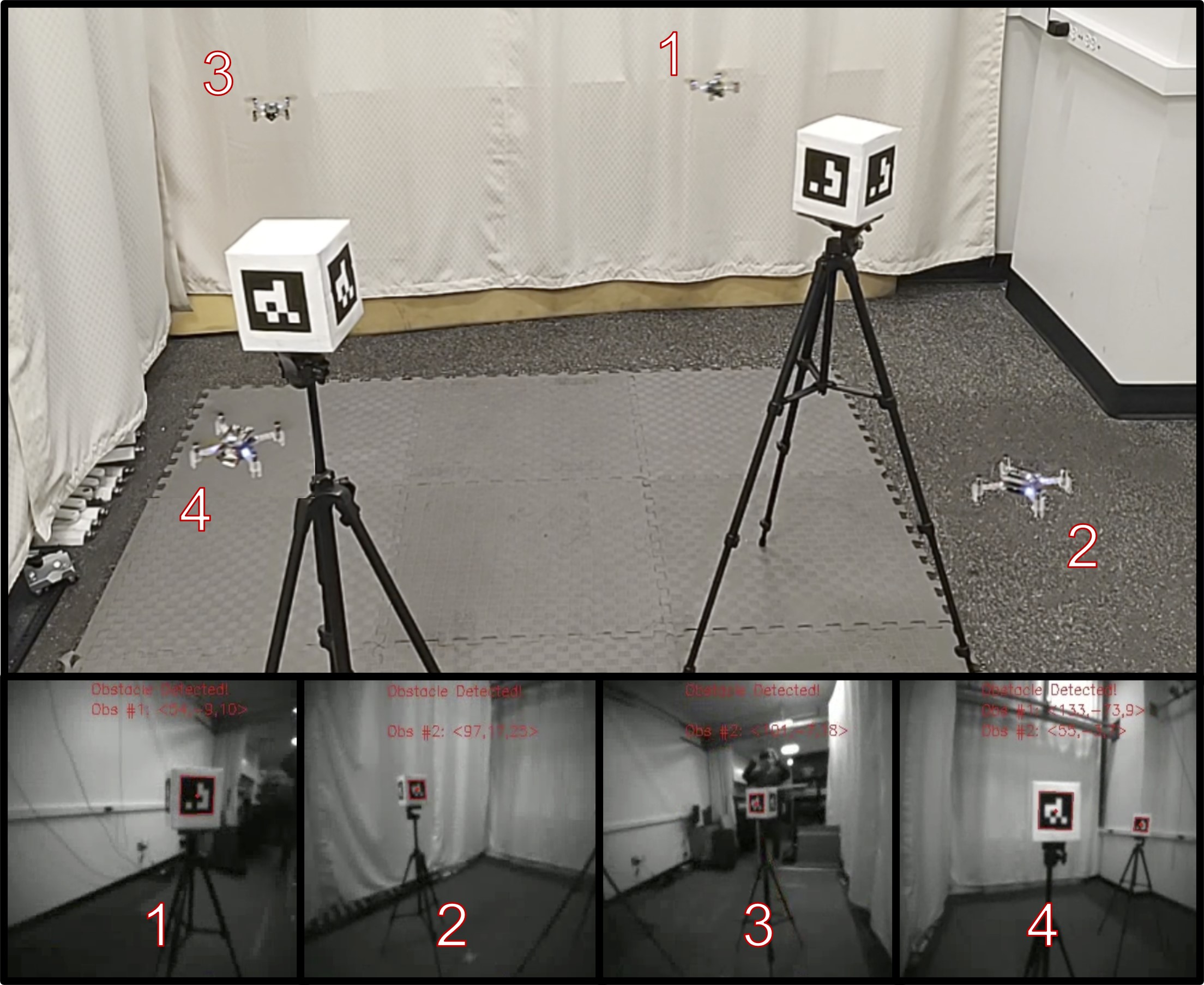}
\caption{\textbf{Flight Demonstration Setup.} The location of the Crazyflie at various times throughout the static collision avoidance demonstration. The corresponding onboard camera image is displayed, highlighting that obstacle information is effectively captured via yaw manipulation. 
\label{ExperimentComposite}}
\end{center}
\vspace{-4ex}
\end{figure}

To demonstrate the experimental flight performance of our safety-aware perception methodology, we implemented it onboard a Crazyflie 2.1 micro-quadrotor UAV. The control architecture, including the safety-critical controller and perception optimization algorithm, was processed in real-time using an onboard STM32F405 \textit{microcontroller unit} (MCU). When solving (\ref{DesiredYaw}), the discrete search increment was set to $9^\text{o}$. Obstacles in the local environment were sensed using a fixed $320\times320$ monochrome optical camera built into the Crazyflie AI-deck. The obstacles were encoded with AprilTag fiducial markers \cite{krogius2019flexible}, as seen in Fig.\;\ref{ExperimentComposite}, for easy and accurate detection, and the associated tag recognition algorithms were processed on a dedicated off-board computer via $2.4\;\text{GHz}$ WiFi communication. The resultant camera frame rate was $7-8\;\text{FPS}$, which was sufficient for static obstacle tracking. The translational state of the UAV was acquired using a Lighthouse Positioning System, and the final control loop was operated at $500\;\text{Hz}$.

\begin{figure}[t]
\begin{center}
\medskip
\includegraphics[width=\linewidth]{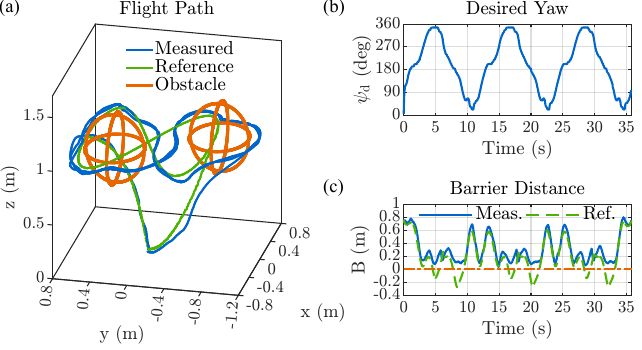}
\caption{\textbf{Flight Demonstration Results.} a) The 3-D flight path of the Crazyflie. The UAV was commanded to follow the position reference (green) while avoiding two static obstacles (orange), resulting in the measured flight trajectory (blue). b) The optimal yaw reference for the Crazyflie. Produced via our online perception algorithm described in Section \ref{Section03}, the UAV attitude controller tracked this reference during the experimental flight. c) The safe-set barrier distance. The distance between the Crazyflie and the closest safe-set boundary (blue) is always greater than zero. Without the safety-critical controller, the UAV would have tracked the reference value (green), confirming that our controller was necessary and sufficient to ensure positive safety margins throughout the flight. 
\label{ExperimentPlot}}
\end{center}
\vspace{-4ex}
\end{figure}

With this experimental setup, we commanded the Crazyflie to track a horizontal infinity symbol trajectory (the same flight described in Mission Profile \#1 in Section\;\ref{Section04}). Two static obstacles were placed in the Crazyflie's flight path, and the UAV was provided with no \textit{a priori} knowledge of the obstacle locations. In order to avoid a collision, the UAV needed to effectively perceive and avoid both obstacles in real-time. A demonstration flight is depicted in Fig.\;\ref{ExperimentComposite}, and the resulting data is presented in Fig.\;\ref{ExperimentPlot}. These results were highly repeatable and representative of overall experimental performance. Exemplary flight videos, which show safety-critical performance in the presence of both static and dynamic obstacles, can be found in the supplemental material.

From Fig.\;\ref{ExperimentComposite} it can be observed that the UAV effectively oriented its onboard camera such that all obstacles were appropriately captured within the FOV. The measured flight path, shown in Fig.\;\ref{ExperimentPlot}(a), closely matched the position reference trajectory when the UAV was unobstructed, and the UAV deviated from the position reference, when necessary, to satisfy all safety-critical constraints. Additionally, the real-time yaw reference angle, plotted in Fig.\;\ref{ExperimentPlot}(b), varied smoothly for optimal collision avoidance. Fig.\;\ref{ExperimentPlot}(c) depicts the measured minimum distance between the Crazyflie and the two obstacle safe-set boundaries. This measured value, which was greater than zero for the entire flight, indicates that the safety-critical requirements were satisfied for the entirety of the demonstration flight. Conversely, without the safety-critical controller the UAV would have tracked the green-dashed reference value and inadvertently collided with both obstacles. 

Finally, for this flight demonstration the average computation time, using the STM32F405 MCU onboard the Crazyflie 2.1, was $371\;\mu\text{s}$. At a control frequency of $500\;\text{Hz}$, this solve-time equates to $19\%$ of the total available solution time, which emphasizes the viability of our approach for online implementation, even on systems with extremely limited computational capabilities. Furthermore, the aforementioned computation time was achieved when solving (\ref{DesiredYaw}) with a discrete angular increment of $9^\text{o}$ which was sufficient for accurate obstacle detection and tracking. The computational complexity of our safety-aware algorithm scales linearly with this increment, enabling easy adjustment of the computation time and resolution to suite the desired application. 

\section{CONCLUSIONS}
\label{Section06}

In this work, we proposed a safety-critical control architecture for autonomous UAVs which incorporates online perception. We detailed our non-invasive approach to dynamic collision avoidance using CBF-based quadratic programming. Next, we presented our novel methodology for safety-aware sensor pointing via the yaw DOF. Inspired by the general sensor coverage control problem, our method utilized a local spatial density function and FOV quality function to maximize environmental risk observations and reduce overall collision risk. To this end, we further employed CBFs to quantify collision risk during obstacle perception and tracking. We initially evaluated performance by conducting an extensive series of simulations using a fixed position reference trajectory and randomly generated dynamic obstacle environments. Our safety-aware algorithm outperformed all other common heuristic approaches by at least $16\%$ with a final safety-critical success rate of $88-96\%$. Lastly, we showed the effectiveness of our online optimization structure in a real-time flight demonstration by running it onboard a Crazyflie UAV. In the presence of these obstacles, the UAV successfully followed its flight trajectory, detected and tracked all obstacles, and maintained safety requirements.

The research presented herein addresses a fundamental challenge for many autonomous systems in dynamical environments, especially in the field of microrobotics. By introducing our safety-aware perception methodology, we lay the foundation for future progress with these computationally-restricted and sensor-limited flight systems. In particular, our algorithm provides a convenient and functional online solution to enable autonomous obstacle detection and dynamic collision avoidance. Our future research is directed towards using this yaw optimization technique, and the associated safety-critical controller, in unique micro-scale applications. We also intend to incorporate a generalized embedded computer vision system, eliminating the need for off-board computing and dedicated positioning systems and paving the road towards complete micro-scale autonomy.

\section*{ACKNOWLEDGMENT}

This research is based upon work supported in part by the Office of the Director of National Intelligence (ODNI), Intelligence Advanced Research Projects Activity (IARPA), via 2021-21090200005. The views and conclusions contained herein are those of the authors and should not be interpreted as necessarily representing the official policies, either expressed or implied, of ODNI, IARPA, or the U.S. Government. The U.S. Government is authorized to reproduce and distribute reprints for governmental purposes notwithstanding any copyright annotation therein.

\bibliographystyle{IEEEtran}
\balance
\bibliography{main}

\end{document}